\documentclass{article}
\PassOptionsToPackage{dvipsnames,table}{xcolor}
\usepackage{arxiv}
\usepackage{graphicx}
\usepackage[numbers]{natbib}
\usepackage{doi}
\usepackage[utf8]{inputenc} 
\usepackage[T1]{fontenc}    
\usepackage{url}            
\usepackage{booktabs}       
\usepackage{amsfonts}       
\usepackage{nicefrac}       
\usepackage{microtype}      
\usepackage{amsmath}
\usepackage{amssymb}
\usepackage{bbm}
\usepackage{wasysym}
\usepackage{pifont}
\usepackage{threeparttable}
\usepackage{subcaption}
\usepackage{array} 
\usepackage{makecell}
\usepackage{algpseudocode} 
\usepackage{multirow}
\usepackage{listings}
\usepackage{enumitem}
\usepackage{hyperref}
\usepackage{todonotes}
\usepackage{comment}
\usepackage[capitalise, noabbrev]{cleveref}
\usepackage{lscape}
\usepackage{tabularray}
\usepackage{CJK}

\usepackage{amsfonts}
\usepackage{amsmath}
\usepackage{multirow}
\usepackage{amssymb}
\usepackage{pifont}
\usepackage{xcolor}
\usepackage{colortbl}
\usepackage{makecell}
\usepackage{wrapfig}

\title{PhysVGGT: Feed-Forward Dense Physical Property Estimation from A Single Image}

\author{Sneha Paul$^{1,2,*}$
\quad Guile Wu$^{1}$
\quad Bingbing Liu$^{3}$
\quad Dongfeng Bai$^{1}$\\
$^{1}$Huawei Noah's Ark Lab
\quad $^{2}$Concordia University
\quad $^{3}$Foundation Model Department, Huawei\\
\texttt{\small{sneha.paul@mail.concordia.ca, \{guile.wu, liu.bingbing, baidongfeng\}@huawei.com}}
}

\renewcommand{\headeright}{Technical Report}

\makeatletter
\g@addto@macro\@maketitle{%
  {\centering
    \includegraphics[width=0.99\linewidth]{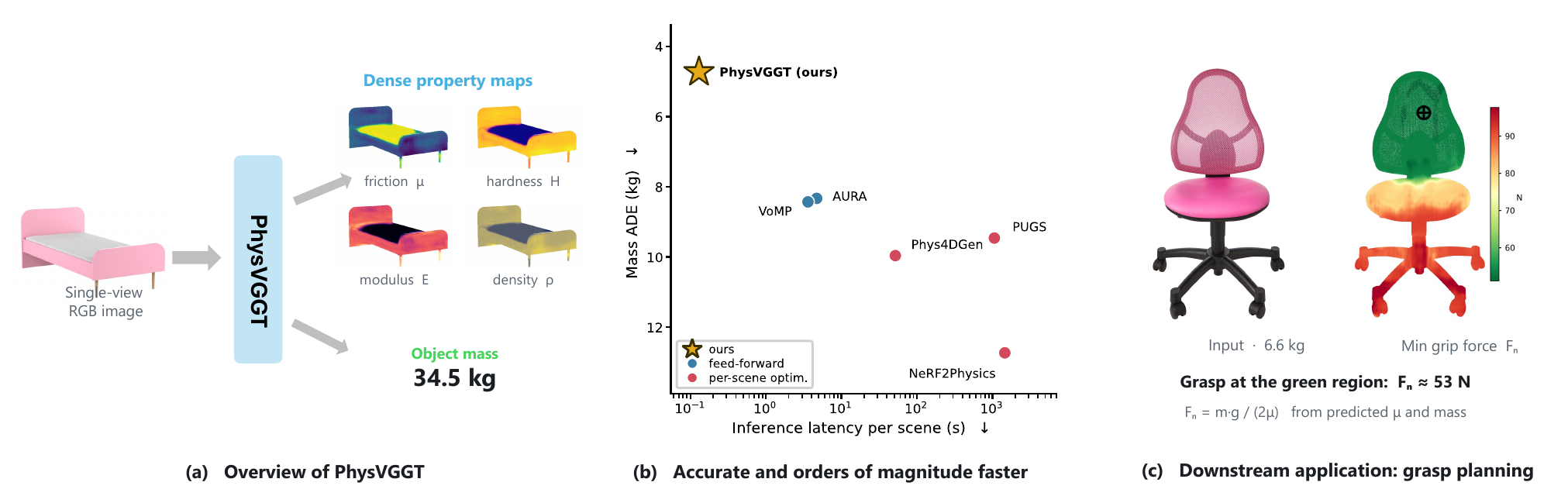}\par
    \captionof{figure}{PhysVGGT overview.
    (a) From a single RGB image, PhysVGGT predicts dense friction $\mu$, hardness $H$, Young's modulus $E$, and density $\rho$, together with object-level mass.
    (b) PhysVGGT achieves accurate physical-property prediction while being $27\times$ faster than the previous state of the art.
    (c) The predicted friction and mass directly enable downstream grasp planning by estimating the minimum grip force $F_n=mg/(2\mu)$ and identifying a low-force contact region.}
    \label{fig:perfeff}\par    
  }
  \vskip 20pt
}
\makeatother

\begin{document}

\maketitle

\begingroup
\renewcommand\thefootnote{*}
\footnotetext{Sneha Paul contributed to this work during an internship at Huawei Canada.}
\endgroup

\thispagestyle{fancy}

\begin{abstract}
Physical properties, such as friction, hardness, stiffness, and density, govern how robots should grasp, manipulate and interact with objects, yet estimating these properties from RGB images remains challenging.
Existing methods typically employ per-object reconstruction augmented with physical properties or directly query vision-language models at test time, which results in substantial computational overhead that limits their applicability.
In this work, we present PhysVGGT, a feed-forward model that predicts dense maps of friction coefficient, Shore hardness, Young's modulus, and density, together with object-level mass, from a single RGB image in one forward pass.
The key idea of PhysVGGT is to formulate physical property estimation as a dense per-pixel prediction problem and employ a visual geometry transformer to extract geometry-aware tokens from the input image followed by a dense prediction branch for estimating local physical properties and a global prediction branch for estimating object-level mass.
In addition, we introduce a scalable pseudo-label generation pipeline that enables large-scale weakly supervised training for dense physical property prediction, substantially reducing the need for expensive direct physical measurements.
Extensive experiments show that PhysVGGT achieves state-of-the-art performance on the ABO-500 dataset and generalizes effectively to the out-of-distribution NeRF2Physics dataset.
Moreover, PhysVGGT eliminates the need for per-object reconstruction and test-time optimization, achieving an inference latency of only $0.13$\,s per image, making it $27\times$ faster than the previous state of the art.
\end{abstract}

\section{Introduction}

Physical properties are latent attributes of the visual world that, unlike geometry or semantics, are difficult to observe directly from images.
Yet humans routinely infer them from visual cues, estimating whether a surface is rough or slippery, whether an object is soft or rigid, or whether it is likely to be light or heavy.
Such reasoning extends visual perception beyond object recognition and geometric understanding toward the physical characteristics that govern how objects and surfaces behave and interact.
Endowing vision systems with this capability is an important step toward richer understanding of physical scenes, with potential benefits for downstream tasks in robotics, autonomous driving, and physics-based applications \citep{xie2024physgaussian,gao2023objectfolder}.

Recent work has demonstrated that physical properties can nevertheless be estimated from visual observations.
One of the most popular paradigms is to leverage Neural Radiance Field (NeRF)~\citep{mildenhall2021nerf} or 3D Gaussian Splatting (3DGS)~\citep{kerbl2023gs} to reconstruct per-object 3D representations from multiple images and then perform semantic or material reasoning to infer physical properties~\citep{zhai2024nerf2physics,xu2025gaussianproperty,shuai2025pugs,chopra2025physgs}.
While effective, this paradigm has significant computational overhead, as it requires expensive reconstruction and inference for each test object.
Some recent works attempt to mitigate this overhead by using more efficient feed-forward approaches that bypass iterative per-object reconstruction.
However, they still face limitations in terms of reliance on prepared 3D assets, test-time Vision-Language Models (VLMs), material retrieval, property lookup, and further post-processing~\citep{dagli2026vomp,pixie,lan2025aura}.
In addition, existing methods typically require \textit{multiple views} of the scene to accurately infer physical properties, which further limits their applicability in real-world scenarios where only a single view may be available.

In this paper, we propose \textbf{PhysVGGT}, a feed-forward model that directly predicts dense physical-property maps of friction coefficient, Shore hardness, Young's modulus, and density, together with object-level mass, from a \emph{single} RGB image in one forward pass.
The key idea of PhysVGGT is to formulate physical property estimation as a dense per-pixel prediction problem and construct a feed-forward framework that directly maps geometry-aware visual features to physical properties.
Specifically, PhysVGGT employs a visual geometry transformer backbone for extracting geometry-aware tokens from the input image, followed by a dense prediction branch for estimating local physical properties and a global prediction branch for estimating object-level mass.
In this framework, PhysVGGT leverages three key components to achieve accurate and efficient physical property estimation.
First, since different physical properties are inherently correlated through the underlying material and geometric characteristics, PhysVGGT introduces a cross-property coupling module into the dense prediction branch, which couples different physical properties at the feature level by attending across property-specific channels at each spatial location.
This design exploits the strong correlations among friction, hardness, stiffness, and density, allowing the prediction of each property to be informed by the others rather than decoded independently.
Second, PhysVGGT introduces a shared decoder with lightweight per-property heads, which replaces independent decoders with a single shared decoding trunk followed by small property-specific output heads.
This design reduces the number of decoder parameters by approximately $4\times$, while the parameter sharing across the correlated prediction tasks provides an effective inductive bias and further improves prediction accuracy.
Third, PhysVGGT introduces an object-level mass head that directly regresses object mass from pooled geometry-aware tokens using a lightweight MLP.
This global prediction pathway is decoupled from the dense decoder, complementing the per-pixel physical-property maps with an object-level scalar estimate.
Together, these designs enable PhysVGGT to train only ${\sim}32\,\mathrm{M}$ parameters on top of a frozen $1.26\,\mathrm{B}$-parameter backbone, while achieving $27\times$ speedup compared to the current state of the art~\citep{dagli2026vomp} and yielding higher prediction accuracy (Figure~\ref{fig:perfeff}).

In addition, we introduce a scalable pseudo-labeling pipeline that generates dense physical-property annotations directly from images, bypassing the need for curated 3D assets and manual physical measurements.
Existing datasets do not provide dense per-pixel annotations of friction, hardness, stiffness, and density, posing a significant challenge for training feed-forward models at scale.
Although there have been some physical property annotation pipelines, they either assign physical attributes to 3D assets, limiting their scalability to the size and diversity of curated asset collections, or infer physical properties through per-object reconstruction and optimization, which is computationally expensive and does not produce a reusable large-scale corpus of annotated images.
In contrast, our pipeline combines part segmentation, material and hardness prediction, and cross-view refinement to generate high-quality pseudo-labels for dense physical-property maps.
This allows us to create a large-scale dataset of densely annotated physical-property maps without the need for direct physical measurements, enabling the training of feed-forward models like PhysVGGT at scale.

We evaluate PhysVGGT on five datasets spanning in-distribution and out-of-distribution settings.
On ABO~\citep{collins2022abo}, we evaluate mass on ABO-500 and qualitatively assess the dense predictions for properties without measured per-pixel ground truth.
In addition, we evaluate zero-shot generalization under a distribution shift to real measurements on NeRF2Physics~\citep{zhai2024nerf2physics} and image2mass~\citep{image2mass}, and qualitatively study terrain generalization on RUGD~\citep{RUGD2019IROS} and RELLIS-3D~\citep{jiang2020rellis3d}.
Our experiments show that PhysVGGT achieves state-of-the-art performance across multiple benchmarks while being faster than existing methods, specifically achieving a $27\times$ inference speedup compared with the previous state of the art.
Our \textbf{contributions} are summarized as follows:

\begin{itemize}
    \item We propose \textbf{PhysVGGT}, a feed-forward framework that directly predicts dense physical properties together with object-level mass from a single RGB image in one forward pass, eliminating per-object reconstruction and test-time optimization.
    \item We conduct extensive experiments and demonstrate that PhysVGGT achieves state-of-the-art performance while running $27\times$ faster than the previous state of the art.
    \item We propose a scalable pseudo-label generation pipeline that enables large-scale weakly supervised training of dense physical property prediction models, eliminating the need for expensive direct physical measurements.
\end{itemize}

\section{Related Work}
\label{sec:related_work}

\paragraph{Physical Properties from Visual Observations.}
Early methods typically build per-object 3D representations and then reason about materials on top of them.
NeRF2Physics~\citep{zhai2024nerf2physics} augments a Neural Radiance Field with language-aligned features to transfer material properties to reconstructed geometry, while GaussianProperty~\citep{xu2025gaussianproperty}, PUGS~\citep{shuai2025pugs} and PhysGS~\citep{chopra2025physgs} attach dense physical attributes to 3D Gaussians through multimodal reasoning, feature propagation and probabilistic fusion.
These methods require per-object reconstruction before any physical reasoning can begin.
Recent methods bypass the need for iterative reconstruction using feed-forward approaches.
AURA~\citep{lan2025aura} replaces iterative reconstruction with a single geometry foundation model pass, yet still queries CLIP, a VLM and a material database at test time, whereas VoMP~\citep{dagli2026vomp} and PIXIE~\citep{pixie} predict volumetric properties from voxel or latent representations produced by a pretrained 3D model.
Phys4DGen~\citep{phys4dgen} embeds physical properties into generative 4D modelling, and \citet{lee2026} regresses global mass from a monocular image under physically guided supervision.
In contrast, PhysVGGT predicts dense friction, Shore hardness, Young's modulus, and density maps, together with object-level mass, from a single RGB image in one forward pass, with neither reconstruction nor any auxiliary semantic model at inference.

\paragraph{Physical Property Data Annotation.}
Since dense physical measurements are labor-intensive to acquire at scale, recent work derives supervision from pretrained models using SAM~2~\citep{ravi2024sam2} for part localization and VLMs combined with material databases for property assignment.
VoMP~\citep{dagli2026vomp} labels the parts of pre-segmented 3D objects with a VLM and a material database and propagates the result to voxels, while PIXIE~\citep{pixie} assembles PixieVerse semi-automatically by distilling a proprietary VLM, CLIP and human priors into per-object material classes.
Both methods are bounded by the scale of curated 3D asset collections and neither covers friction or Shore hardness.
On the other hand, per-scene pipelines such as NeRF2Physics~\citep{zhai2024nerf2physics} and GaussianProperty~\citep{xu2025gaussianproperty} instead query GPT-4/GPT-4V for one reconstructed object at a time and therefore never yield a reusable training corpus.
We follow the weak-supervision direction, but with an offline pipeline that operates purely on multiview product imagery, needs no 3D assets, and produces dense per-pixel supervision at an order of magnitude more objects.

\begin{figure}[t!]
\centering
\includegraphics[width=0.9\linewidth]{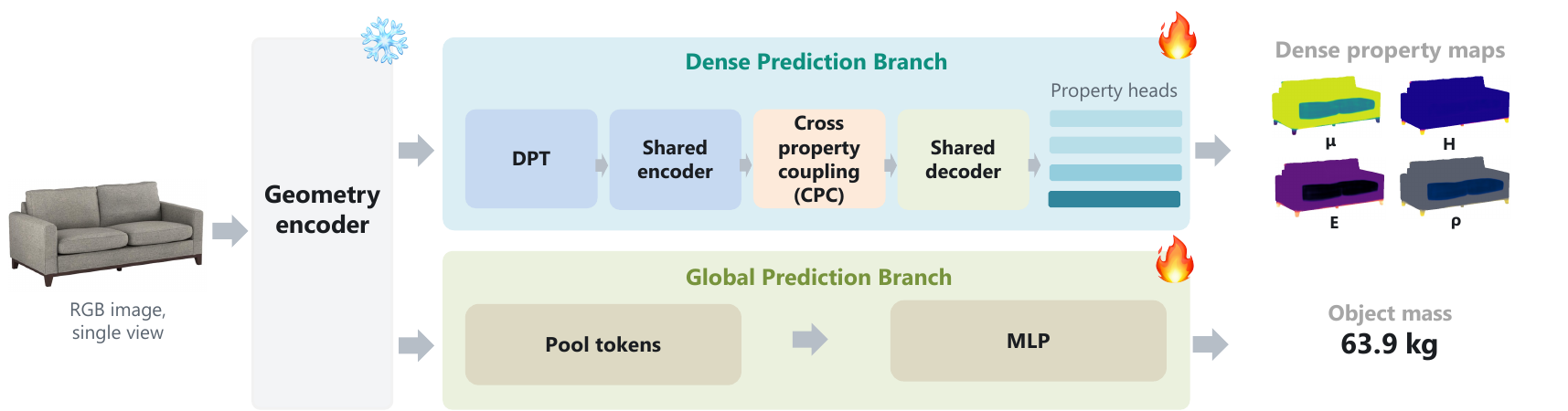}
\caption{The overall framework of PhysVGGT.} 
\label{fig:model}
\end{figure}

\section{Method}

\paragraph{Problem Statement.}
In this work, PhysVGGT focuses on directly predicting dense physical properties and object-level mass from a single RGB image in a feed-forward manner, rather than relying on 3D reconstruction or iterative optimization.
Let \(I\in\mathbb{R}^{3\times H\times W}\) denote a single RGB image of an object.
We consider four dense physical properties \(\mathcal{C}=\{\mu,H,E,\rho\}\), corresponding to the coefficient of friction, Shore hardness, Young's modulus and density.
For each property \(c\in\mathcal{C}\) the goal is to predict a dense property map \(\widehat{Y}^{c}\in[0,1]^{H\times W}\), together with an object-level mass estimate \(\widehat{M}\in\mathbb{R}_{>0}\).
PhysVGGT therefore formulates dense physical property estimation as a feed-forward prediction problem \((\widehat{\mathbf{Y}},\widehat{M})=f_{\Theta}(I)\), where \(\widehat{\mathbf{Y}}=\{\widehat{Y}^{c}\mid c\in\mathcal{C}\}\) is the set of predicted dense property maps and \(f_{\Theta}\) the proposed network.

\subsection{Feed-Forward Physical Property Estimation: PhysVGGT}

\paragraph{Geometry Encoder.}
Visual geometry transformers~\citep{wang2025vggt,wang2025pi3,zhuo2025streamvggt} have shown strong capabilities in capturing 3D spatial structure from 2D images and been demonstrated well-suited for serving as backbones in scene understanding tasks~\citep{wu20254dlangvggt}.
Inspired by these advances, we adopt a visual geometry transformer as the backbone (geometry encoder $\Phi_{\mathrm{geo}}$) to extract geometry-aware token representations \(T\) from the input RGB image \(I\), which serve as the foundation for subsequent physical property prediction:
\begin{equation}
    T=\Phi_{\mathrm{geo}}(I).
\end{equation}
Then, as shown in Figure~\ref{fig:model}, PhysVGGT predicts the dense physical property maps \(\widehat{\mathbf{Y}}\) and the object-level mass \(\widehat{M}\) through two parallel prediction branches.

\paragraph{Dense Prediction Branch.}
We first convert the geometry tokens into a dense spatial representation using a DPT-style feature extractor~\citep{ranftl2021dpt}.
Multi-scale token representations from multiple transformer stages are projected into a common feature space and hierarchically fused through a top-down refinement pathway to produce a high-resolution dense feature representation:
\begin{equation}
    F
    =
    \mathcal{H}_{\mathrm{DPT}}(T),
\end{equation}
where \(F\in\mathbb{R}^{d_f\times H\times W}\).
The dense feature map is subsequently processed by a shared encoder that progressively aggregates local and global context while reducing the spatial resolution:
\begin{equation}
    (S_1,\ldots,S_L,B)
    =
    \mathcal{E}_{\theta}(F),
\end{equation}
where \(S_l\) denotes the skip feature at the \(l\)th resolution and \(B\in\mathbb{R}^{d_b\times H_b\times W_b}\) is the bottleneck feature map containing the highest-level semantic representation.
Although each physical property is predicted independently by a separate head, these properties are governed by shared underlying material characteristics and exhibit strong physical correlations.
To explicitly model these dependencies, we introduce a \textit{Cross-Property Coupling (CPC)} module that operates on the shared bottleneck representation and performs self-attention across property-specific latent representations.
For each property \(c\in\mathcal{C}\), the shared bottleneck feature is projected into a property-specific embedding:
\begin{equation}
    z^{(0)}_{c,x}
    =
    A_c(B(x))
    +
    e_c,
\end{equation}
where \(A_c\) is a learnable projection, \(e_c\) is a learnable property embedding, and \(x\) denotes a spatial location.
The embeddings are grouped as \(Z_x^{(0)}=[z^{(0)}_{c,x}]_{c\in\mathcal{C}}\) and refined by a lightweight transformer, \(Z_x=\mathcal{T}(Z_x^{(0)})\), which enables information exchange among correlated physical properties at each spatial location.
The coupled representations are projected back to the shared feature space through residual connections:
\begin{equation}
    B_c(x)
    =
    B(x)
    +
    O_c(Z_{x,c}),
\end{equation}
where \(O_c\) denotes a learnable output projection.
The output projections are initialized to zero so that the CPC module initially behaves as an identity mapping and gradually learns residual cross-property interactions during training.
Each property-specific bottleneck representation is independently processed by a shared decoder that progressively restores the spatial resolution using the corresponding encoder skip features:
\begin{equation}
    U_c
    =
    \mathcal{D}_{\phi}(B_c,S_1,\ldots,S_L),
\end{equation}
and followed by a lightweight property-specific prediction head:
\begin{equation}
    \widehat{Y}^{c}
    =
    \sigma\!\left(h_c(U_c)\right),
\end{equation}
where \(h_c\) is the property-specific prediction layer and \(\sigma(\cdot)\) denotes the sigmoid activation.
Sharing the feature extractor, encoder, and decoder enables efficient multi-task learning while allowing the CPC module to capture dependencies among correlated physical properties and the prediction heads to preserve property-specific calibration.

\paragraph{Global Prediction Branch.}
In parallel, object-level mass is predicted directly from the extracted geometry tokens.
Since object mass depends primarily on global object geometry rather than local material characteristics, the geometry tokens are globally pooled, batch-normalized per feature to preserve the magnitude cue encoding object scale and processed by an MLP:
\begin{equation}
    \bar{t}=\operatorname{Pool}(T),\qquad
    \widehat{\ell}_M=g_{\psi}(\bar{t}),\qquad
    \widehat{M}=\exp(\widehat{\ell}_M),
\end{equation}
where \(\widehat{\ell}_M\) is the predicted logarithmic mass.
Unlike previous approaches that estimate mass from predicted density and reconstructed volume, we directly regress mass from the global geometry-aware representation.
This avoids error accumulation from intermediate density and volume estimation while allowing the mass branch to specialize independently from the dense prediction branch.

\subsection{Training Objective}

PhysVGGT is jointly optimized for dense physical-property prediction and object-level mass estimation.
Since the dense supervision is automatically generated (Section~\ref{sec:pseudo_label}) and may contain local noise or calibration errors, we combine a robust reconstruction objective with a pairwise ranking loss that emphasizes reliable relative relationships.
Let $Y^c$ and $\widehat{Y}^c$ denote the pseudo ground-truth and predicted maps for property $c\in\mathcal{C}$, respectively.
Given the foreground mask $M_{\mathrm{fg}}$, we define the valid pixels as $\Omega_{\mathrm{fg}}=\{x\mid M_{\mathrm{fg}}(x)=1\}$ and apply the robust Charbonnier penalty~\citep{charbonnier1994}:
\begin{equation}
    \mathcal{L}^{\mathrm{rec}}_c
    =
    \frac{1}{|\Omega_{\mathrm{fg}}|}
    \sum_{x\in\Omega_{\mathrm{fg}}}
    \rho\!\left(\widehat{Y}^{c}(x)-Y^{c}(x)\right).
\end{equation}
Absolute pseudo-label values can be noisy even when their relative ordering is reliable.
We therefore complement reconstruction with a ranking objective that preserves the ordering of physical properties across regions.
To avoid ambiguous supervision, we retain only pixel pairs whose pseudo-label difference exceeds a threshold:
\begin{equation}
    \mathcal{Q}_c =
    \left\{(i,j)\in\Omega_{\mathrm{fg}}^2 :
    |Y^c(i)-Y^c(j)|>\varepsilon_{\mathrm{rank}}\right\}.
\end{equation}
For each retained pair, we define $s_{ij}^c=\operatorname{sign}(Y^c(i)-Y^c(j))$ and minimize
\begin{equation}
    \mathcal{L}^{\mathrm{rank}}_c
    =
    \frac{1}{|\mathcal{Q}_c|}
    \sum_{(i,j)\in\mathcal{Q}_c}
    \operatorname{softplus}\!\left(
    -\frac{
    s_{ij}^c\bigl(\widehat{Y}^c(i)-\widehat{Y}^c(j)\bigr)-\gamma
    }{\tau}
    \right),
\end{equation}
where $\tau$ is the temperature and $\gamma$ is the ranking margin.
We sample $2048$ foreground pairs per image and property and use $\tau=0.1$, $\gamma=0$, and $\varepsilon_{\mathrm{rank}}=0.02$.
Since the properties are normalized to $[0,1]$, the threshold removes pairs differing by less than $2\%$ of the prediction range.
The dense objective combines reconstruction and ranking across all properties:
\begin{equation}
    \mathcal{L}_{\mathrm{dense}}
    =
    \sum_{c\in\mathcal{C}}
    \lambda_c
    \left(
    \mathcal{L}^{\mathrm{rec}}_c
    +
    \lambda_{\mathrm{rank}}\mathcal{L}^{\mathrm{rank}}_c
    \right),
\end{equation}
where $\lambda_c$ balances individual properties and
$\lambda_{\mathrm{rank}}$ controls the ranking contribution.
For object-level mass, we regress log-mass using an $\ell_1$ loss:
\begin{equation}
    \mathcal{L}_{\mathrm{mass}}
    =
    \left|\widehat{\ell}_M-\log M\right|,
\end{equation}
where $M$ is the ground-truth object mass. Samples without valid mass annotations contribute only to the dense objective.
The final training objective is
\begin{equation}
    \mathcal{L}
    =
    \mathcal{L}_{\mathrm{dense}}
    +
    \lambda_{\mathrm{mass}}\mathcal{L}_{\mathrm{mass}}.
\end{equation}

\subsection{Pseudo-Label Generation Pipeline}
\label{sec:pseudo_label}

\begin{figure}[t]
\centering
\includegraphics[width=0.8\linewidth]{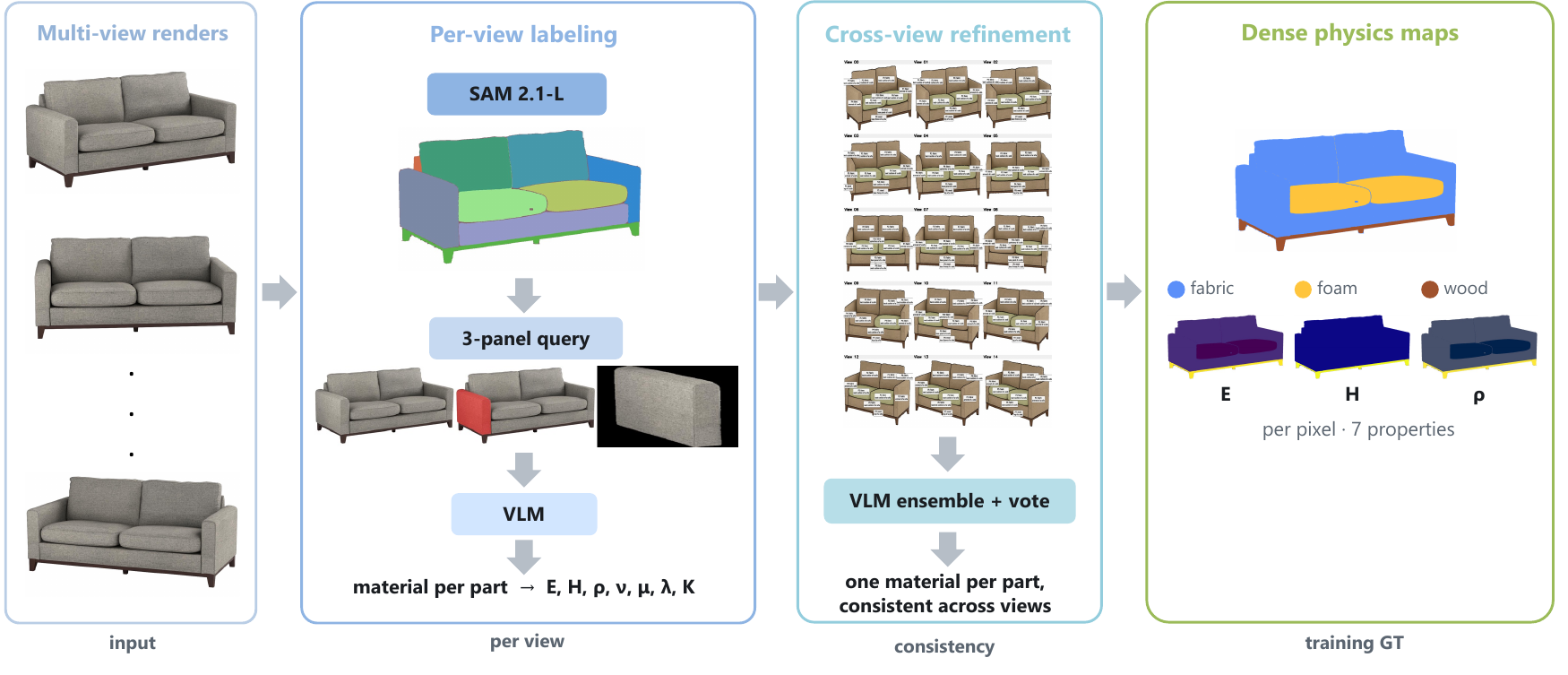}
\caption{The overview of the proposed pseudo-label generation pipeline.}
\label{fig:datagen}
\end{figure}

Since large-scale datasets with dense physical-property annotations are unavailable, we construct an offline pseudo-label generation pipeline (Figure~\ref{fig:datagen}) to produce supervision for PhysVGGT.
Given multi-view images \(\mathcal{I}\) of an object, the set of generated dense physical property maps $\mathbf{Y}=\{Y^{c}\mid c\in\mathcal{C}\}$ is obtained by:
\begin{equation}
    \mathbf{Y}
    =
    \mathcal{G}(\mathcal{I}),
\end{equation}
where \(\mathcal{G}\) denotes the pseudo-label generation pipeline that works in three stages: part segmentation, material and hardness prediction, and cross-view refinement.
\textbf{(I) Part segmentation.} Each view is decomposed into part-level regions with SAM~2~\citep{ravi2024sam2} to divide image into parts.
This is important since different parts may contain different material and have different physical properties. 
\textbf{(II) Material and hardness prediction.} For every part we construct a three-panel query image, including the full object, the object with the part highlighted, and a tight crop of the part, and prompt a VLM to return a material category from a fixed 12-material vocabulary together with a Shore hardness range and its scale (Shore~A or Shore~D).
The three-panel format supplies global object context and local appearance detail simultaneously, and predictions below a confidence threshold are discarded.
\textbf{(III) Cross-view refinement.} Since the views are labelled independently, the same physical part may be assigned different materials in different views.
All views are therefore tiled into a single composite image and presented to the VLM with a prompt that asks it to locate and correct the inconsistent parts; the corrections are combined by majority vote over \((\text{view},\text{part})\) pairs, yielding a single material per part that is consistent across all views.

The physical properties are then derived from the refined labels.
Young's modulus is computed from the predicted Shore hardness through an empirical hardness-modulus relation and clipped to the literature range of the assigned material, while density, Poisson's ratio and the coefficient of friction are read from a material-property table compiled from standard engineering references.
The Lam\'e parameters and the bulk modulus follow in closed form from \(E\) and \(\nu\).
Finally, the per-part values are rasterized onto the image plane to produce the dense supervision \(\mathbf{Y}\).
Object-level mass annotations are obtained independently from the ABO product metadata.
The resulting training dataset is
\begin{equation}
    \mathcal{D}
    =
    \left\{
    \left(
    \mathcal{I},
    \mathbf{Y},
    M
    \right)
    \right\},
\end{equation}
where \(M\) denotes the object-level mass annotation.

\begin{table}[!htb]
\centering
\caption{{Mass estimation on ABO-500} (100 test objects).
Baselines quoted from the VoMP protocol \citep{dagli2026vomp} and the respective papers.
$\downarrow$/$\uparrow$ denote lower/higher is better.
Best in \textbf{bold}.
$^\star$Taken from existing work for reference purpose.
}
\label{tab:mass}
\footnotesize
\setlength{\tabcolsep}{5pt}
\begin{tabular}{lcccl}
\toprule
Method & ADE (kg)\,$\downarrow$ & ALDE\,$\downarrow$ & MnRE\,$\uparrow$ & Paradigm \\
\midrule
NeRF2Physics \citep{zhai2024nerf2physics} & 12.73 & 0.736 & 0.564 & per-scene NeRF + VLM \\
PUGS \citep{shuai2025pugs}                &  9.46 & 0.661 & 0.576 & per-scene 3DGS \\
Phys4DGen$^\star$ \citep{phys4dgen}       &  9.96 & 0.664 & 0.566 & 3DGS + MLLM \\
VoMP \citep{dagli2026vomp}                &  8.43 & 0.631 & 0.576 & feed-forward \\
PGVME \citep{lee2026}                &  8.32 & 0.998 & --    & feed-forward \\
AURA \citep{lan2025aura}                  &  8.33 & 0.582 & 0.634 & feed-forward + VLM \\
PhysGS$^\star$ \citep{chopra2025physgs}    &  8.25 & 0.999 & 0.474 & per-scene 3DGS + VLM \\
GaussianProperty \citep{xu2025gaussianproperty} & 5.96 & 0.744 & 0.559 & 3DGS + MLLM \\
\midrule
\textbf{PhysVGGT (ours)} & \textbf{4.73} & \textbf{0.574} & \textbf{0.655} & feed-forward \\
\bottomrule
\end{tabular}
\end{table}

\begin{table}[!htb]
\centering
\caption{{Friction and Shore hardness on the NeRF2Physics 13-scene real
benchmark}.
Reproduced baselines under the protocol of \citet{zhai2024nerf2physics}.}
\label{tab:fh}
\footnotesize
\setlength{\tabcolsep}{3pt}
\begin{tabular}{lcccccccc}
\toprule
& \multicolumn{4}{c}{\textbf{Friction} $\mu$} & \multicolumn{4}{c}{\textbf{Shore hardness}} \\
\cmidrule(lr){2-5}\cmidrule(lr){6-9}
Method & ADE\,$\downarrow$ & ALDE\,$\downarrow$ & APE\,$\downarrow$ & MnRE\,$\uparrow$
       & ADE\,$\downarrow$ & ALDE\,$\downarrow$ & APE\,$\downarrow$ & MnRE\,$\uparrow$ \\
\midrule
NeRF2Physics \citep{zhai2024nerf2physics} & 0.217 & 0.459 & 0.446 & 0.661 & 66.21 & 0.672 & 1.105 & 0.546 \\
PhysGS \citep{chopra2025physgs}           & 0.218 & 0.462 & 0.438 & 0.647 & 34.82 & 0.456 & 0.532 & 0.671 \\
GPT-4V \citep{achiam2023gpt}                      & 0.209 & 0.430 & 0.549 & 0.692 & 32.75 & 0.330 & 0.304 & 0.758 \\
CLIP \citep{radford2021clip}              & 0.222 & 0.455 & 0.602 & 0.654 & 32.86 & 0.294 & 0.266 & 0.774 \\
\midrule
\textbf{PhysVGGT (ours)} & \textbf{0.102} & \textbf{0.206} & \textbf{0.225} & \textbf{0.831}
                         & \textbf{7.05}  & \textbf{0.103} & \textbf{0.111} & \textbf{0.908} \\
\bottomrule
\end{tabular}
\end{table}

\section{Experiments}
\label{sec:experiments}


\textbf{Datasets.}
We evaluate PhysVGGT on two benchmarks used by prior work on physical property estimation.
\emph{(i)~Mass:} the ABO-500 test split \citep{zhai2024nerf2physics} (100 held-out objects with catalogue-measured ground-truth weight).
\emph{(ii)~Friction \& hardness:} the NeRF2Physics 13-scene real dataset \citep{zhai2024nerf2physics}, which provides sparse in-the-wild measurements of the kinetic friction coefficient ($\mu$) and Shore hardness on real objects; this is an out-of-distribution (OOD) test, disjoint from our ABO training data.
For qualitative out-of-distribution generalization we additionally use the off-road semantic-segmentation datasets RUGD \citep{RUGD2019IROS} and RELLIS-3D \citep{jiang2020rellis3d}.
We train PhysVGGT on ABO-8k, a set of 8{,}213 household objects drawn from the Amazon Berkeley Objects catalogue \citep{collins2022abo}, each rendered from 15 calibrated views ($\sim$123k images), with per-view dense property maps. 
Object-level \emph{mass} uses the catalogue item weight, available for 7{,}326 of the objects.
The model is trained directly on these maps without any per-scene optimization.
To prevent leakage, the \textit{100 ABO-500 test objects are held out} from all training.

\textbf{Metrics.}
Following \citep{zhai2024nerf2physics,dagli2026vomp}, we report the Absolute Distance Error (ADE, in physical units), the Absolute Log-Distance Error $\mathrm{ALDE}=|\log \hat{y}-\log y|$, the Absolute Percentage Error $\mathrm{APE}=|\hat{y}-y|/y$, and the Min-of-Ratio Error $\mathrm{MnRE}=\min(\hat{y}/y,\,y/\hat{y})\in(0,1]$.

\begin{table}[!t]
\centering
\caption{{Inference efficiency.}
Per-item inference latency.
Baseline latencies are reported from \citep{dagli2026vomp,lan2025aura} and the corresponding papers.
Lower is better.}
\label{tab:eff}
\footnotesize
\setlength{\tabcolsep}{6pt}
\begin{tabular}{lcc}
\toprule
Method & Latency (s)\,$\downarrow$ & Test-time dependency \\
\midrule
NeRF2Physics \citep{zhai2024nerf2physics} & 1454.6 & NeRF optimization + VLM \\
PUGS \citep{shuai2025pugs}        & 1058.3 & 3DGS optimization \\
Pixie \citep{pixie}      & 201.6  & Scene optimization \\
Phys4DGen \citep{phys4dgen} & 51.7 & 3DGS + MLLM \\
AURA \citep{lan2025aura}        & 4.7  & CLIP + VLM \\
VoMP \citep{dagli2026vomp}        & 3.6  & DINOv2 + voxelization \\
\midrule
\textbf{PhysVGGT (ours)} & \textbf{0.13} & \textbf{None} \\
\bottomrule
\end{tabular}
\end{table}

\begin{figure}[!htb]
\centering
\includegraphics[width=0.70\linewidth]{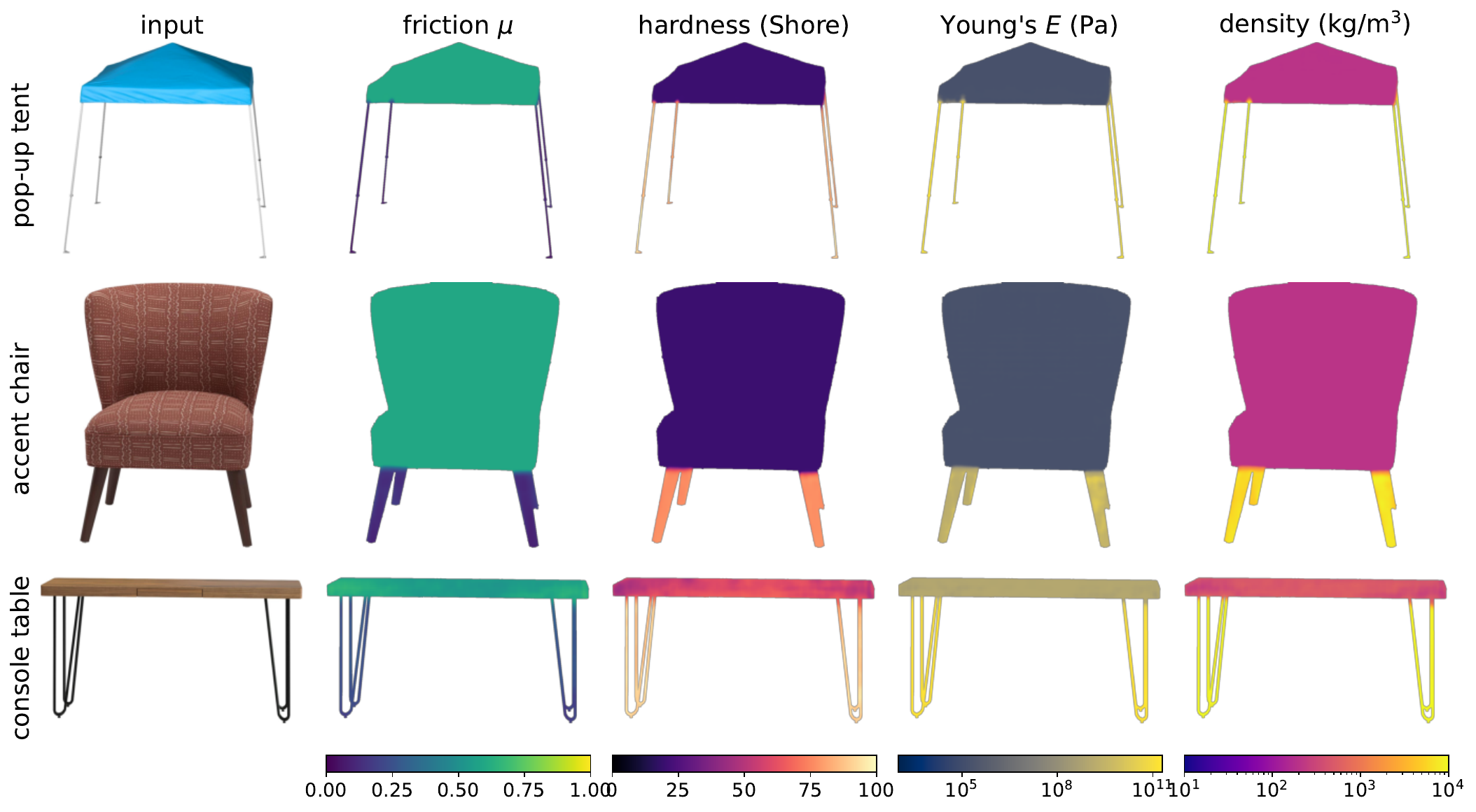}
\caption{{Dense physical property predictions.}
From the single input view (left), PhysVGGT predicts per-pixel friction $\mu$, Shore hardness, Young's modulus $E$ and density in one forward pass. 
}
\label{fig:heatmaps}
\end{figure}

\begin{table}[t]
\centering
\caption{{Component-wise ablation.}
N2P MnRE on a fixed $2000$-object subset using identical training settings,
averaged over four epochs.}
\label{tab:components}
\small
\begin{tabular}{lcc}
\toprule
Configuration & Friction MnRE\,$\uparrow$ & Hardness MnRE\,$\uparrow$ \\
\midrule
\textbf{Full} (trunk + coupling + shared dec. + combined loss)
& \textbf{0.831} & \textbf{0.908} \\
\ \ $-$ w/o cross-property coupling
& 0.805 & 0.869 \\
\ \ $-$ w/o shared decoder ($\to$ 4 independent)
& 0.810 & 0.860 \\
\ \ $-$ w/o combined loss ($\to$ plain L1)
& 0.750 & 0.784 \\
\bottomrule
\end{tabular}
\end{table}

\subsection{Comparison with State-of-the-Arts}
\label{sec:main}

\textbf{Mass.}
Table~\ref{tab:mass} reports object-level mass estimation on ABO-500.
PhysVGGT improves on the best previously reported value of every metric: it lowers ADE by $21\%$ relative to GaussianProperty, the strongest prior result on that metric, and by $43\%$ relative to AURA, the strongest prior method on ALDE and MnRE, while also improving ranking consistency, and it outperforms NeRF- and 3DGS-based pipelines despite using a single RGB image at inference.
Directly regressing mass from geometry-aware representations is thus a more accurate and robust alternative to reconstruction- or density-based estimation.

\textbf{Friction and Hardness.}
Table~\ref{tab:fh} reports zero-shot transfer to the out-of-distribution NeRF2Physics real benchmark.
PhysVGGT is best on every metric for both properties, reducing friction ADE by 51.2\% (MnRE $+20.1$\%) and Shore hardness ADE by 78.5\% (MnRE $+17.3$\%), outperforming both reconstruction-based and vision-language baselines on unseen real objects.

\textbf{Efficiency.}
PhysVGGT infers in a single feed-forward pass with no per-scene optimization and no test-time VLM, requiring only 0.13\,s per image, a \(27\times\) reduction over VoMP~\citep{dagli2026vomp}, the fastest existing feed-forward method (Table~\ref{tab:eff}).
Figure~\ref{fig:perfeff} plots this accuracy-latency trade-off against prior work.
The proposed physics head contains only \(32.4\)M trainable parameters on top of a frozen StreamVGGT backbone and requires just \(4.1\) TFLOPs per image.

\subsection{Qualitative Analysis}
\label{sec:qualitative}
Figure~\ref{fig:heatmaps} shows dense predictions for all four properties on multi-material objects.
No benchmark provides \emph{measured}-$E$ or density ground truth, so a head-to-head comparison is not possible, but the maps are physically consistent and material-aware: the model separates the tent's fabric canopy from its metal poles, the chair's upholstery from its wooden legs, and the table's wooden top from its steel hairpin legs, assigning the metal parts high stiffness and density but low friction.

\begin{table}[t]
\centering
\caption{{Cross-property coupling ablation.}
Comparison of different feature interaction mechanisms for modeling
dependencies among physical properties. 
}
\label{tab:cpc}
\small
\setlength{\tabcolsep}{6pt}
\begin{tabular}{lccc}
\toprule
Coupling & Friction & Hardness & Mass \\
         & MnRE\,$\uparrow$ & ADE\,$\downarrow$ & MnRE\,$\uparrow$ \\
\midrule
w/o CPC          & 0.802 & \phantom{0}9.09 & 0.474 \\
Attention, 1-layer      & 0.822 & 10.79 & 0.468 \\
\textbf{Attention, 2-layer (default)} & \textbf{0.831} & \textbf{7.05} & \textbf{0.655} \\
Cross-stitch     & 0.801 & 10.16 & 0.488 \\
FiLM                    & 0.790 & 13.50 & 0.469 \\
\bottomrule
\end{tabular}
\end{table}

\begin{figure}[t]
\centering
\includegraphics[width=0.75\linewidth]{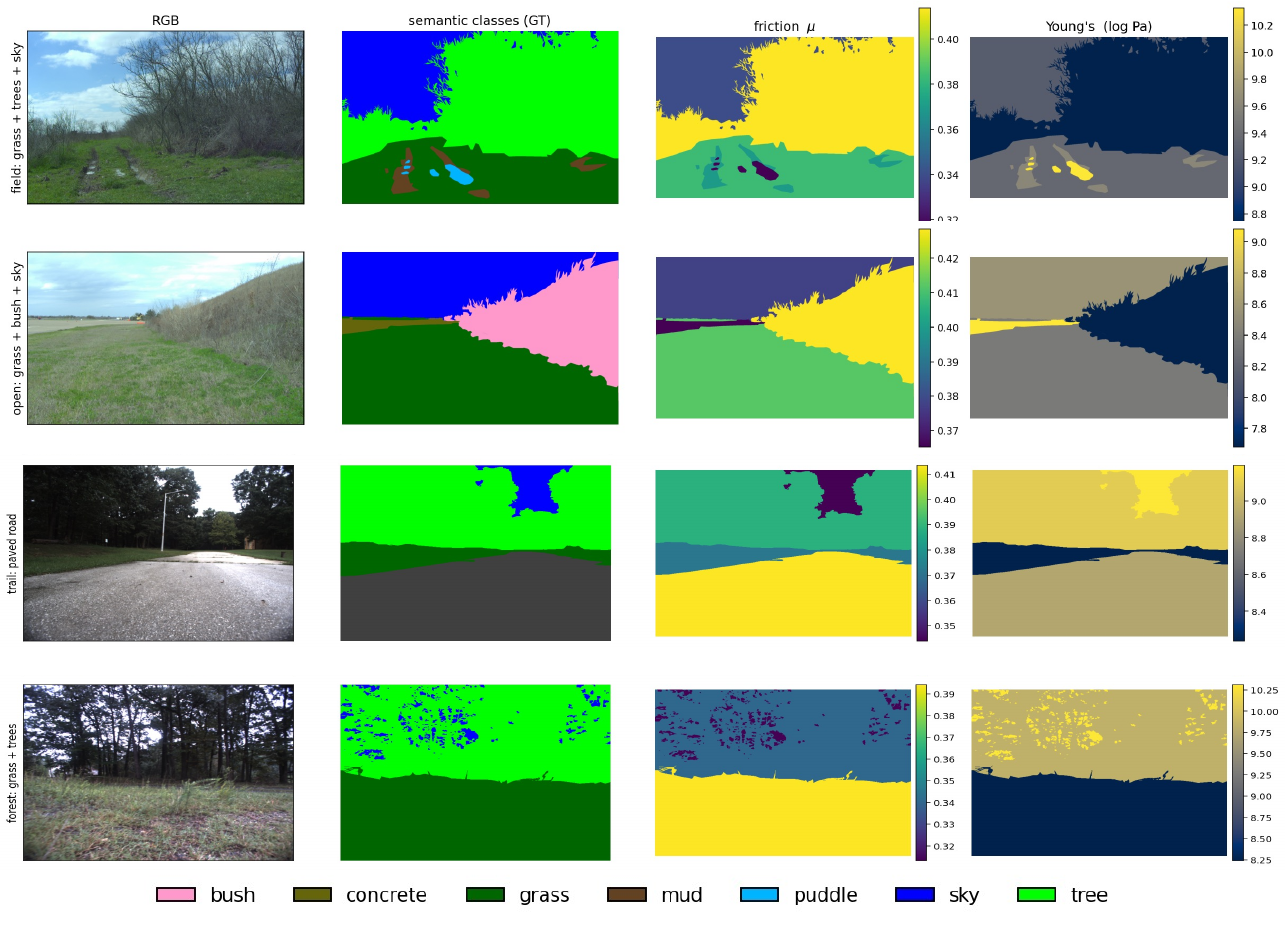}
\caption{{Out-of-distribution scenes} on RUGD (upper two examples) and RELLIS-3D (lower two examples).
Ground-truth semantic regions coloured by our predicted friction and Young's modulus.
The model differentiates unseen materials.}
\label{fig:ood}
\end{figure}

\subsection{Ablation Study and Discussion}

\textbf{Component-Wise Ablation Study.}
We compare the \emph{Full} model against leave-one-out variants on the same $2000$-object subset with identical hyperparameters, reporting N2P MnRE averaged over four epochs.
Table~\ref{tab:components} shows that the combined loss is the main contributor: replacing it with L1 reduces friction 
and hardness by $9.7\%$ and $13.7\%$, respectively.
In contrast, removing cross-property coupling costs $3.1\%$ friction and $4.3\%$ hardness, and removing the shared decoder $2.5\%$ and $5.3\%$, both far smaller than the loss.
Overall, the Full model provides the best performance.

\textbf{Cross-Property Coupling.}
Table~\ref{tab:cpc} compares strategies for modeling interactions among physical properties.
Against the no-CPC variant, the proposed two-layer attention module improves friction MnRE by 3.6\%, reduces hardness ADE by
22.4\% and improves mass MnRE by 38.2\%.
Cross-stitch networks~\citep{misra2016crossstitch} and FiLM~\citep{perez2018film} capture only limited interactions, confirming that attending across friction, hardness, Young's modulus and density is what matters.

\textbf{Domain Generalization Capability.}
\label{sec:zeroshot}
Generalization of PhysVGGT holds beyond object-centric imagery.
Figure~\ref{fig:ood} shows dense predictions on the off-road RUGD~\citep{RUGD2019IROS} and RELLIS-3D~\citep{jiang2020rellis3d} scenes: despite the shift from household objects to natural terrain, the predicted friction and Young's modulus still differentiate materials never seen during training.

\textbf{Please refer to the Appendix for more ablation study and hyperparameter sensitivity analysis.}

\section{Conclusion}
This work presents PhysVGGT, a feed-forward framework that estimates dense physical properties, including friction, Shore hardness, Young's modulus, and density, together with object-level mass, from a single RGB image.
PhysVGGT consists of a geometry encoder that extracts appearance-aware 3D representations from a single RGB image, followed by a dense prediction branch that estimates dense physical properties and a global prediction branch that estimate object-level mass.
Extensive experiments show that PhysVGGT outperforms state-of-the-art methods in terms of both prediction accuracy and inference efficiency.

{
\bibliographystyle{plainnat}
\bibliography{ref}
}

\appendix
\section{Appendix}

In this section, we present more details on the implementation and discuss additional experimental results.
The following is an overview of the organization of the appendix.

\begin{description}[noitemsep, topsep=4pt, parsep=0pt, partopsep=0pt]
    \item[\ref{sec:hparams}] Implementation Details
    \item[\ref{sec:limitation_future_work}] Limitation and Future Work
    \item[\ref{sec:arch}] Architectural Design
    \item[\ref{sec:ablations}] More Ablation Study
    \item[\ref{sec:sens}] Hyperparameter Sensitivity
    \item[\ref{sec:discussion}] Further Discussion and Analysis
\end{description}

\begin{table}[t]
\centering
\caption{{Hyperparameters of the deployed PhysVGGT.}
The frozen StreamVGGT backbone (1.26\,B parameters) receives no gradient;
all listed settings apply to the 32.4\,M-parameter physics heads.}
\label{tab:hparams}
\small
\begin{tabular}{ll}
\toprule
\multicolumn{2}{l}{\emph{Architecture}} \\
\midrule
backbone (frozen)            & StreamVGGT, 1.26\,B params \\
backbone token dim $C$       & 2048 \\
DPT feature width            & 128 \\
UNet head width $h$          & 64 \ ($h,2h,4h,8h$ stages) \\
dense properties             & 4 \ ($\mu$, $H$, $E$, $\rho$) \\
coupling                     & attention, 2 layers \\
coupling dim / heads         & 128 / 4 \\
decoder                      & shared trunk $+$ per-property heads \\
decoder dropout              & 0.1 \\
trainable parameters         & 32.4\,M \\
\midrule
\multicolumn{2}{l}{\emph{Optimization}} \\
\midrule
optimizer                    & AdamW, $\beta=(0.9,0.95)$ \\
learning rate                & $4\times10^{-5}$, constant \\
weight decay                 & $1\times10^{-4}$ \\
warm-start LR multiplier     & 0.1 \\
batch size                   & 4 per GPU $\times$ 4 GPUs \\
epochs                       & 8 \\
gradient clipping            & 1.0 \\
seed                         & 42 \\
input mode / resolution      & centre crop, $518$ \\
\midrule
\multicolumn{2}{l}{\emph{Losses}} \\
\midrule
reconstruction               & Charbonnier, foreground-masked \\
$\lambda_{\mu}$ / $\lambda_{H}$ / $\lambda_{E}$ / $\lambda_{\rho}$ & 3 / 1 / 1 / 1 \\
$\lambda_{\mathrm{rank}}$, $\tau$ & 0.25, 0.1 \\
$\lambda_{\mathrm{mass}}$    & 1.0 \ (log-space) \\
\bottomrule
\end{tabular}
\end{table}

\subsection{Implementation Details}
\label{sec:hparams}

We implement our approach with Python and PyTorch.
We initialize the geometry encoder backbone with pretrained weights from~\cite{zhuo2025streamvggt}, freeze all backbone parameters ($\sim$1.26B), and train only the proposed shared encoder-decoder, CPC module and physics heads ($\sim$32.4M).
Models are optimized using AdamW~\citep{loshchilov2019adamw} with \((\beta_1,\beta_2)=(0.9,0.95)\), a learning rate of \(4\times10^{-5}\), weight decay \(10^{-4}\), batch size \(4\), and gradient clipping with a maximum norm of \(1.0\). 
During training, multiple calibrated RGB views are jointly processed by the frozen backbone, whereas inference requires only a single RGB image.
All experiments are conducted on the same single data-center GPU system with approximately 140 GB of high-bandwidth memory under identical hardware setting.
Table~\ref{tab:hparams} lists the complete configuration of the deployed model.

\subsection{Limitation and Future Work}
\label{sec:limitation_future_work}
Although PhysVGGT demonstrates strong performance in estimating physical properties from a single image, an inherent limitation is that it relies solely on visual information, which can be misleading in cases where the appearance does not accurately reflect the underlying physical properties.
Incorporating additional sensory modalities could complement visual cues, potentially improving the robustness and accuracy of physical property estimation.
Future work will focus on addressing this limitation and integrating PhysVGGT into robotic systems to evaluate its impact on robotic perception and manipulation.

\subsection{Architectural Design}
\label{sec:arch}

\begin{figure}[!t]
\centering
\includegraphics[width=1\linewidth]{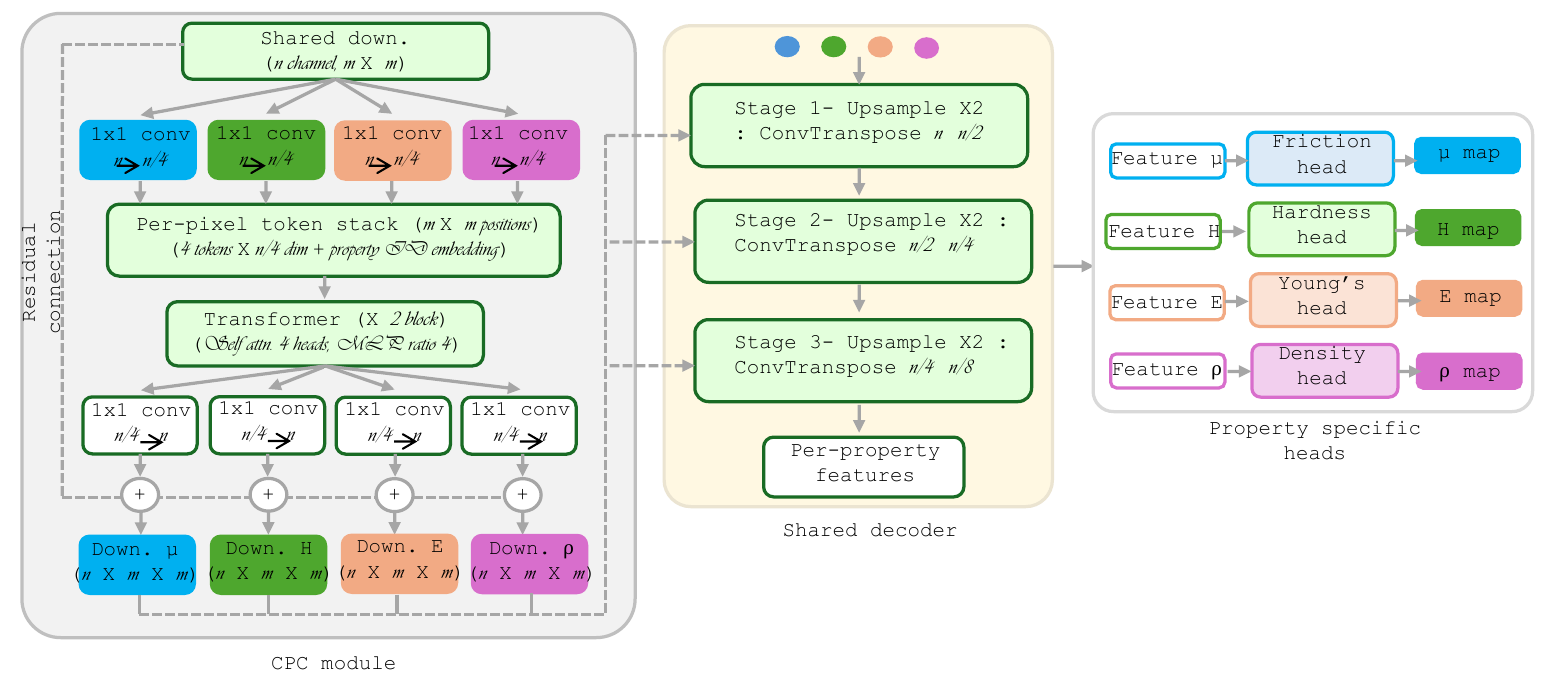}
\caption{{Detailed architectural design of PhysVGGT's cross-property coupling (CPC) module, shared decoder, and property specific head.} }
\label{fig:physvggt}
\end{figure}

PhysVGGT attaches lightweight trainable heads to the frozen geometry encoder backbone.
The trainable modules include the DPT feature extractor, UNet encoder, CPC module, shared
decoder, and mass head, which have $22.8$M, $5.1$M, $0.9$M, $3.0$M, and $0.55$M
parameters, respectively.
In total, these trainable modules account for approximately $32.4$M parameters.
The detailed architectural design of cross-property coupling (CPC) module, shared decoder, and property specific head is illustrated in Figure \ref{fig:physvggt}.

\paragraph{Cross-Property Coupling.}
Physical properties are correlated, yet independent output heads cannot explicitly exchange information. CPC introduces this interaction at the encoder bottleneck, where features are most semantic and spatial resolution is lowest.
The $8h=512$-channel bottleneck is projected into four $128$-D property tokens at each spatial location. Learned property embeddings identify the tokens, and two pre-norm transformer blocks (4 attention heads, MLP ratio 4) model cross-property dependencies. Property-specific projections then map the tokens back to $512$ channels and are added residually to the bottleneck.
Attention operates across the four \emph{properties}, rather than spatial locations, making its cost independent of image resolution. CPC therefore adds only $0.9$M parameters ($2.8\%$ of the head). We zero-initialize the output projections, making CPC an identity mapping at initialization and enabling stable initialization from an uncoupled checkpoint.

\paragraph{Shared Decoder.}
Frozen backbone tokens are converted to dense features by a DPT head and passed through a four-stage UNet encoder with widths $(h,2h,4h,8h)$, where $h=64$.
Rather than using an independent decoder for each property, we employ a single shared upsampling trunk with encoder skip connections, followed by lightweight property-specific $3\times3$ output heads.
This design exploits the common spatial structure of physical properties, which typically share material boundaries while differing in their values. Bilinear interpolation before skip fusion makes the decoder resolution-agnostic.
Sharing improves both efficiency and accuracy: compared with four independent decoders, it reduces the head from $40.9$M to $31.9$M parameters while improving friction MnRE from $0.808$ to $0.831$ and hardness ADE from $8.62$ to $7.05$.
Thus, sharing provides a useful inductive bias in addition to parameter savings.

\paragraph{Physics Property and Mass Heads.}
The shared decoder predicts four dense properties: friction, Shore hardness, Young's modulus, and density.
Friction is predicted directly, hardness is normalized to $[0,1]$, and Young's modulus and density are predicted in $\log_{10}$ space to accommodate their large dynamic ranges.
We optimize these maps using a foreground-masked Charbonnier reconstruction loss together with a pairwise ranking loss.
Mass is treated separately because it is an object-level rather than per-pixel quantity.
Instead of deriving mass from the dense bottleneck, the mass branch operates directly on pooled frozen backbone tokens, which empirically improves MnRE from $0.55$ to $0.61$.
Specifically, we average the patch tokens from the last aggregator layer and apply an MLP $2048\!\rightarrow\!256\!\rightarrow\!64\!\rightarrow\!1$ with BatchNorm, GELU, and dropout, predicting $\log m$.
This branch runs in parallel with the dense pathway, so mass prediction does not alter the dense features.
BatchNorm is preferred over LayerNorm because it preserves sample-level magnitude information useful for estimating object scale, improving MnRE from $0.547$ to $0.605$.
Finally, predicting mass in log space better accommodates its large dynamic range and aligns training with the ratio-based evaluation metrics.

\begin{table}[t]
\centering
\caption{{Spatial dense trunk vs.\ pooled-token regression}.
The per-pixel trunk is essential for the spatial properties.}
\label{tab:dense_ablation}
\small
\setlength{\tabcolsep}{6pt}
\begin{tabular}{lcccc}
\toprule
& \multicolumn{2}{c}{Friction $\mu$} & \multicolumn{2}{c}{Shore hardness} \\
\cmidrule(lr){2-3}\cmidrule(lr){4-5}
Head & MnRE\,$\uparrow$ & ADE\,$\downarrow$ & MnRE\,$\uparrow$ & ADE\,$\downarrow$ \\
\midrule
Pooled-token & 0.742 & 0.148 & 0.844 & 12.52 \\
\textbf{Dense trunk (ours)}  & \textbf{0.831} & \textbf{0.102} & \textbf{0.908} & \textbf{7.05} \\
\bottomrule
\end{tabular}
\end{table}

\subsection{Ablation and Sensitivity Study}
\label{sec:ablations}

\paragraph{Dense Prediction vs.\ Pooled Tokens.}
Table~\ref{tab:dense_ablation} compares direct regression from pooled backbone tokens against the proposed shared dense trunk. 
The dense design improves friction MnRE by 12.0\% (ADE $-31.1$\%) and hardness MnRE by 7.6\% (ADE $-43.6$\%): friction and hardness are spatial properties that depend on local material variation and benefit from pixel-wise prediction.
This motivates a shared dense trunk for the four properties, with object-level mass predicted separately from globally pooled geometry features.

\begin{table}[t]
\centering
\caption{{Choice of geometry backbone} under identical head and training protocol.}
\label{tab:backbone}
\small
\begin{tabular}{lcccc}
\toprule
& \multicolumn{2}{c}{Friction $\mu$} & \multicolumn{2}{c}{Shore hardness} \\
\cmidrule(lr){2-3}\cmidrule(lr){4-5}
Backbone & MnRE\,$\uparrow$ & ALDE\,$\downarrow$ & MnRE\,$\uparrow$  & ALDE\,$\downarrow$ \\
\midrule
\textbf{StreamVGGT (ours)} & \textbf{0.831} & \textbf{0.206} & \textbf{0.908} &  \textbf{0.103 }\\
VGGT-1B \citep{wang2025vggt}       & 0.819 & 0.292 & 0.875 & 0.159 \\
DINOv2-L \citep{oquab2024dinov2}    & 0.802 & 0.252 & 0.834 & 0.173 \\
\bottomrule
\end{tabular}
\end{table}

\paragraph{Choice of Geometry Backbone.}
Table~\ref{tab:backbone} compares different frozen backbone representations while keeping the prediction head and training protocol unchanged.
StreamVGGT and VGGT perform comparably, however, the streaming formulation preserves the geometry representation needed for single-image estimation. 
DINOv2 trails on both friction and hardness, while StreamVGGT is best on both — the geometry-aware VGGT-family features benefit both properties.

\begin{table}[t!]
\centering
\caption{{Backbone freezing strategy.}
Freezing the entire StreamVGGT backbone vs.\ fine-tuning its global-attention
blocks.}
\label{tab:freeze}
\small
\setlength{\tabcolsep}{6pt}
\begin{tabular}{lcccc}
\toprule
StreamVGGT & Train. & Friction & Hardness & Mass \\
freezing   & params & MnRE\,$\uparrow$ & MnRE\,$\uparrow$ & MnRE\,$\uparrow$ \\
\midrule
\textbf{Fully frozen (ours)} & \textbf{32\,M} & \textbf{0.831} & \textbf{0.908} & \textbf{0.655} \\
Unfreeze global-attn   & 334\,M         & 0.695 & 0.814 & 0.517 \\
\bottomrule
\end{tabular}
\end{table}

\begin{table}[t!]
\centering
\caption{{Decoder ablation.} Decoder architectures for multi-task
physical property prediction.}
\label{tab:decoder}
\small
\setlength{\tabcolsep}{5pt}
\begin{tabular}{lccccc}
\toprule
Decoder & Dec. & Total & Fric. & Hard. & Hard. \\
        & params & train. & MnRE\,$\uparrow$ & ADE\,$\downarrow$ & MnRE\,$\uparrow$ \\
\midrule
original (4 decoders) & 12.1\,M & 40.9\,M & 0.808 & 8.62 & 0.887 \\
slim (4 half-width)   &  4.3\,M & 33.2\,M & 0.808 & 7.48 & 0.899 \\
\textbf{shared + heads (ours)} & \textbf{3.0\,M} & \textbf{31.9\,M} & \textbf{0.831} & \textbf{7.05} & \textbf{0.908} \\
\bottomrule
\end{tabular}
\end{table}

\paragraph{Backbone Freezing Strategy.}
Table~\ref{tab:freeze} evaluates whether fine-tuning the backbone helps.
Unfreezing the global-attention blocks raises the trainable parameter count from 32\,M to 334\,M yet degrades every task, and the fully frozen model improves friction, hardness and mass MnRE by 19.6\%, 11.5\% and 26.7\% respectively.
Fine-tuning under weak pseudo supervision likely overfits the pretrained representation, so optimizing only the lightweight physics head gives both higher accuracy and far better parameter efficiency.

\paragraph{Shared Decoder.}
Table~\ref{tab:decoder} compares decoder designs.
Replacing four independent decoders with a single shared decoder plus lightweight property-specific heads cuts decoder parameters by 75.2\% and total trainable parameters by 22.0\%, yet still improves friction MnRE by 2.8\% and reduces hardness ADE by 18.2\%.
The four properties evidently share a common spatial representation, so parameter sharing aids both efficiency and generalization.

\begin{table}[t!]
\centering
\caption{{Loss-component ablation}.
}
\label{tab:loss}
\small
\setlength{\tabcolsep}{6pt}
\begin{tabular}{lccc}
\toprule
Loss & Friction & Hardness & Mass \\
     & MnRE\,$\uparrow$ & MnRE\,$\uparrow$ & ADE\,$\downarrow$ \\
\midrule
L1 baseline                  & 0.684 & 0.826 & 12.39 \\
\ + foreground mask          & 0.759 & 0.856 & 10.72 \\
\ + pairwise ranking         & 0.811 & 0.851 & 11.49 \\
\ + all (mask, Charb., rank) & \textbf{0.831} & \textbf{0.908} & \textbf{4.73} \\
\bottomrule
\end{tabular}
\end{table}

\begin{table}[t!]
\centering
\caption{{Data scaling.} Mass prediction on ABO-500 and OOD friction/hardness on N2P. Performance improves consistently with training-set size.}
\label{tab:scaling}
\footnotesize
\setlength{\tabcolsep}{5pt}
\begin{tabular}{lccc cc}
\toprule
& \multicolumn{3}{c}{Mass (ABO-500)} & \multicolumn{2}{c}{OOD (N2P)} \\
\cmidrule(lr){2-4}\cmidrule(lr){5-6}
Train.\ obj. & ADE (kg)\,$\downarrow$ & ALDE\,$\downarrow$ & MnRE\,$\uparrow$
             & $\mu$ MnRE\,$\uparrow$ & $H$ MnRE\,$\uparrow$ \\
\midrule
\phantom{0}500  & 5.62 & 0.633 & 0.651 & 0.787 & 0.824 \\
2000            & 5.26 & 0.658 & 0.641 & 0.794 & 0.860 \\
7225            & \textbf{4.73} & \textbf{0.574} & \textbf{0.655} & 
 \textbf{0.831} & \textbf{0.908} \\
\bottomrule
\end{tabular}
\end{table}

\begin{table}[t!]
\centering
\caption{{Mass branch: dense ($\rho\!\times\!V$) versus global (direct
head)} on ABO-500. Both geometry tokens come from the same frozen geometry backbone and the
same trained model; only the path from features to mass differs.}
\label{tab:massbranch}
\small
\begin{tabular}{lcccc}
\toprule
Mass branch & ADE (kg)\,$\downarrow$ & ALDE\,$\downarrow$ & APE\,$\downarrow$ & MnRE\,$\uparrow$ \\
\midrule
dense, $\bar{\rho}\!\times\!V_{\text{shell}}\!\times\!\kappa$ & 10.3 & 0.896 & 1.043 & 0.517 \\
global, direct mass head & \textbf{4.73} & \textbf{0.574} & \textbf{0.787} & \textbf{0.655} \\
\bottomrule
\end{tabular}
\end{table}

\begin{table}[t!]
\centering
\caption{{Refined vs.\ raw supervision} comparison.}
\label{tab:raw_refined_gt}
\small
\setlength{\tabcolsep}{4pt}
\begin{tabular}{lccc}
\toprule
Supervision & Friction & Hardness & Mass (dense) \\
\midrule
Raw & 0.813 & 0.856 & 0.450 \\  
Refined & \textbf{0.831} & \textbf{0.908} & \textbf{0.655} \\
\bottomrule
\end{tabular}
\end{table}

\paragraph{Loss Components.}
We ablate foreground masking, Charbonnier reconstruction, and pairwise ranking losses under controlled 3-epoch runs (Table~\ref{tab:loss}).
Ranking primarily improves friction, increasing MnRE by $18.6\%$, while foreground masking improves hardness by $3.6\%$.
Combining all components achieves the best overall performance, improving friction and hardness MnRE by $21.5\%$ and $9.9\%$ and reducing mass ADE by $61.8\%$ over the L1 baseline, and is used by default.

\paragraph{Data Scaling.}
We study the effect of training-set size using $500$, $2000$, and $7225$ objects (Table~\ref{tab:scaling}).
Going from $500$ to $7225$ training objects lowers mass ADE on ABO-500 by $15.8\%$, although mass MnRE is essentially flat ($+0.6\%$) and dips at $2000$ objects.
The out-of-distribution N2P properties benefit more clearly and monotonically: friction MnRE improves by $5.6\%$ and hardness MnRE by $10.2\%$.

\paragraph{Dense vs.\ Global Mass Prediction.}
\label{sec:massbranch}
We compare two strategies for mass prediction: deriving mass from dense density and volume following the existing SOTA methods $m=\bar{\rho}V\kappa$, and directly regressing $\log m$ from pooled backbone tokens.
As shown in Table~\ref{tab:massbranch}, direct regression performs substantially better, improving MnRE from $0.517$ to $0.655$ and reducing ADE from $10.3$ to $4.73$\,kg.
The dense route is limited by coarse material-density labels, volume-estimation errors, and error propagation through $\rho V$.
In contrast, the global head directly optimizes the target quantity and leverages all $7326$ ABO product weights without requiring volume annotations.
We therefore use the global branch for mass prediction and retain dense heads for spatial physical properties.

\begin{table}[t!]
\centering
\caption{{Hyperparameter sensitivity.} N2P MnRE ($\uparrow$) under
one-at-a-time sweeps from the final configuration ($\star$).}
\label{tab:sens}
\small
\begin{tabular}{llcc}
\toprule
Hyperparameter & Setting & $\mu$ MnRE\,$\uparrow$ & $H$ MnRE\,$\uparrow$ \\
\midrule
\multirow{4}{*}{head width $h$}
  & 32          & 0.792 & 0.855 \\
  & 48          & \textbf{0.839} & \textbf{0.877} \\
  & 64\,$\star$ & 0.829 & \textbf{0.877} \\
  & 96          & 0.822 & 0.859 \\
\midrule
\multirow{4}{*}{$\lambda_{\mathrm{friction}}$}
  & 1           & 0.807 & 0.864 \\
  & 2           & 0.770 & 0.868 \\
  & 3\,$\star$  & 0.829 & \textbf{0.877} \\
  & 5           & 0.799 & 0.859 \\
\midrule
\multirow{4}{*}{learning rate}
  & $2\times10^{-5}$ & 0.837 & 0.864 \\
  & $4\times10^{-5}\,\star$ & 0.829 & \textbf{0.877} \\
  & $8\times10^{-5}$ & 0.770 & 0.846 \\
  & $1.6\times10^{-4}$ & \multicolumn{2}{c}{\emph{diverged (NaN)}} \\
\midrule
\multirow{4}{*}{$\lambda_{\mathrm{rank}}$}
  & 0             & 0.741 & 0.845 \\
  & 0.1           & 0.781 & 0.863 \\
  & 0.25\,$\star$ & 0.829 & \textbf{0.877} \\
  & 0.5           & \textbf{0.878} & 0.865 \\
\midrule
\multirow{4}{*}{seed}
  & 42\,$\star$ & 0.829 & 0.877 \\
  & 1           & 0.847 & 0.865 \\
  & 2           & 0.798 & 0.854 \\
  & 3           & 0.767 & 0.853 \\
\bottomrule
\end{tabular}
\end{table}

\begin{table}[t!]
\centering
\caption{{Zero-shot mass prediction on image2mass.}
All baselines are trained on image2mass~\citep{image2mass}, whereas PhysVGGT is trained only on ABO and evaluated without fine-tuning.
$\dagger$: human estimates provided with the dataset.}
\label{tab:i2m}
\footnotesize
\begin{tabular}{lccc}
\toprule
Method & Zero-shot & ALDE\,$\downarrow$  & MnRE\,$\uparrow$ \\
\midrule
Xception $k$-NN            & No & 1.454 &  0.341 \\
Xception $k$-NN (SIM)      & No & 0.781 & 0.534 \\
Pure CNN                   & No & 0.680 &  0.575 \\
CNN with ALDE              & No & 0.666 &  0.573 \\
No Geometry                & No & 0.520 &  0.638 \\
Shape-aware                & No & 0.465 &  0.675 \\
Shape-aware, multi-view    & No & \textbf{0.437} &  \textbf{0.693} \\
\midrule
\textbf{PhysVGGT (ours)}   & Yes & 0.585 &  0.653 \\
\midrule
\emph{Human median}$^{\dagger}$   & --- & 0.459 &  0.675 \\
\emph{Human ensemble}$^{\dagger}$ & --- & 0.323 & 0.754 \\
\bottomrule
\end{tabular}
\end{table}

\paragraph{Refine vs.\ Raw Pseudo Label Supervision.}
\label{sec:raw_refined_gt}
Table~\ref{tab:raw_refined_gt} compares training on the raw per-view pseudo labels with training on the cross-view refined labels.
Refinement improves every property: friction MnRE rises from $0.813$ to $0.831$ ($+2.2\%$), hardness from $0.856$ to $0.908$ ($+6.1\%$), and dense mass from $0.450$ to $0.655$ ($+45.6\%$).
The effect is smallest on friction and by far the largest on dense mass, which aggregates predictions over the whole object and is therefore the most exposed to a part being assigned different materials in different views.
This confirms the value of the cross-view refinement stage in the pseudo-label pipeline.

\subsection{Hyperparameter Sensitivity}
\label{sec:sens}

We assess sensitivity to head width, friction-loss weight, learning rate, and ranking-loss weight by varying one hyperparameter at a time from the final configuration ($\star$).
All runs use the same $2000$-object subset for 2 epochs; we additionally evaluate 4 random seeds to estimate training variance. Thus, Table~\ref{tab:sens} is intended to compare relative trends rather than absolute performance.

Table~\ref{tab:sens} shows that PhysVGGT is generally robust to hyperparameter choices, with most variations comparable to the observed seed variance.
The ranking loss has the clearest effect: removing it decreases friction MnRE from $0.829$ to $0.741$ ($-10.6\%$), while increasing its weight to $0.5$ reaches $0.878$.
Head capacity largely saturates at $h=48$--$64$, and moderate changes to $\lambda_{\mathrm{friction}}$ have limited impact.
The learning rate remains stable up to $4\times10^{-5}$, degrades at $8\times10^{-5}$, and diverges at $1.6\times10^{-4}$.
We retain $\lambda_{\mathrm{rank}}=0.25$ despite the higher friction score at $0.5$ because the final configuration was fixed before this sweep, avoiding tuning on the evaluation benchmark.

\subsection{Further Analysis and Discussion}
\label{sec:discussion}

\begin{figure}[t]
\centering
\includegraphics[width=0.96\linewidth]{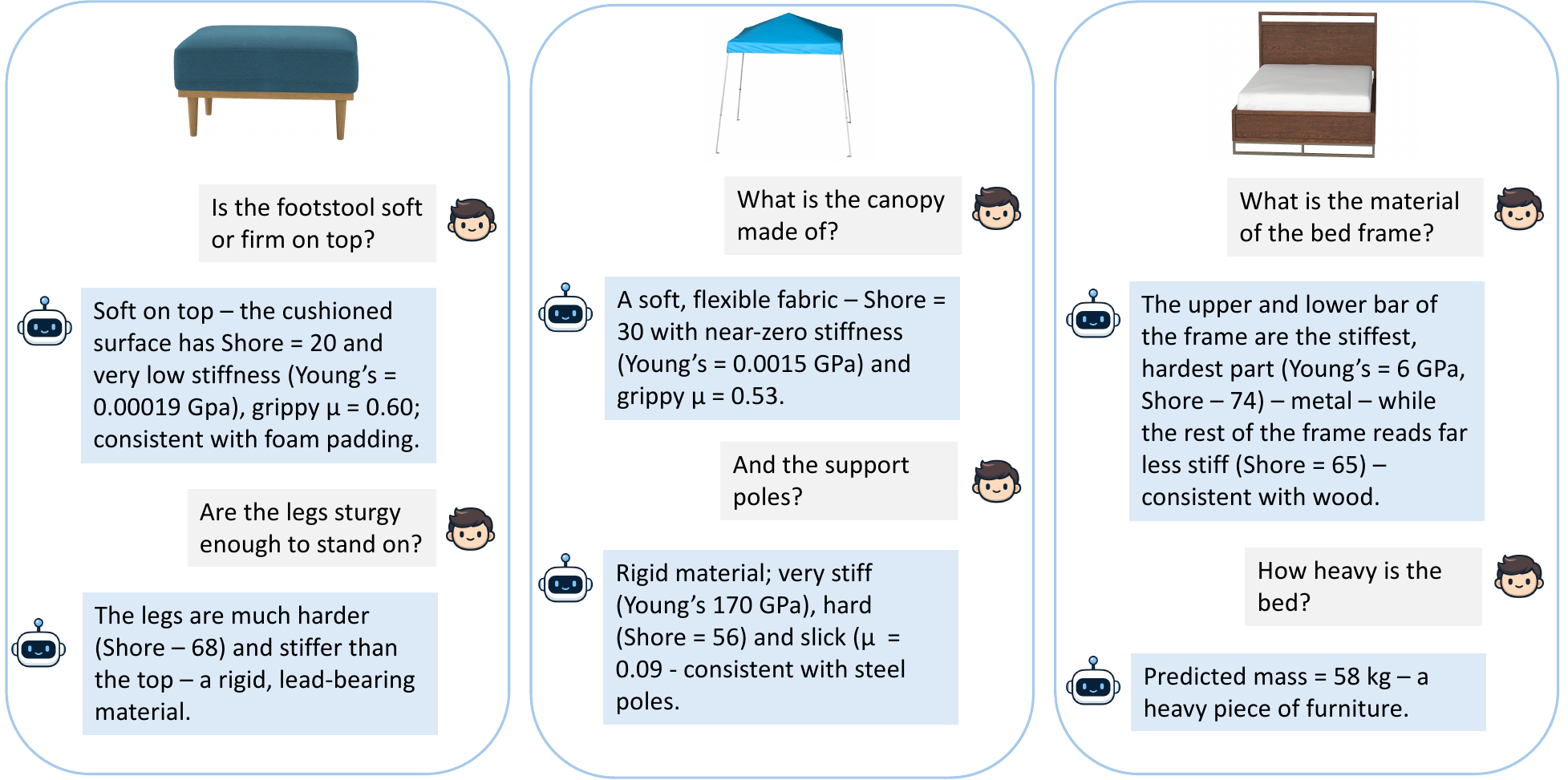}\\[2pt]
\includegraphics[width=0.85\linewidth]{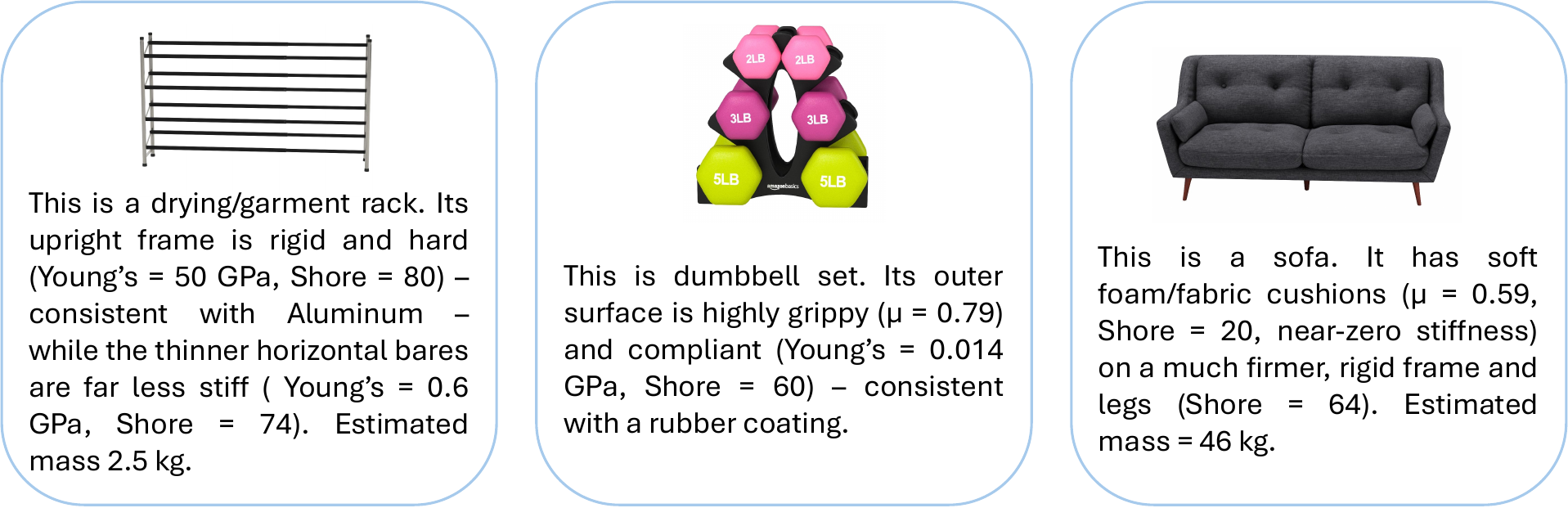}
\caption{{Language-guided physical reasoning.}
\emph{Top:} multi-round question answering about object- and region-level physical properties.
\emph{Bottom:} single-shot physical-composition captioning.
PhysVGGT supplies dense friction, Shore hardness, Young's modulus, density, and object-level mass predictions; SAM2 supplies spatial regions; and Qwen3-VL interprets the original and segmented images and generates the responses.
Material names are estimated by matching each predicted property tuple $(\mu,H,E,\rho)$ to a reference material database.
All numerical physical claims are derived from PhysVGGT predictions.}
\label{fig:lang}
\end{figure}

\paragraph{PhysVGGT Supports Grounded, Interpretable Physical Reasoning.}
Figure~\ref{fig:lang} demonstrates a language-guided interface built on PhysVGGT predictions.
Given a single RGB image, PhysVGGT predicts dense physical properties and object-level mass, while SAM2 defines spatial regions.
The original image, segmented image, and region-level predictions are provided to Qwen3-VL for multi-round question answering and physical-composition captioning.
Thus, visual semantics and language generation come from the VLM, whereas all reported physical quantities are grounded in PhysVGGT predictions.

\paragraph{Domain Generalization Capability.}
We further evaluate the ABO-trained mass head on the real-world image2mass household benchmark~\citep{image2mass} without fine-tuning.
This represents a substantial domain shift from synthetic ABO renders to real images of everyday household objects.
As shown in Table~\ref{tab:i2m}, PhysVGGT achieves $0.653$ MnRE, outperforming five of seven baselines despite all baselines being trained in-domain on approximately $150$k image2mass products.
It trails only the shape-aware models ($0.675$/$0.693$), which additionally use explicit object dimensions, and is within $0.022$ MnRE of the median human estimate ($0.675$).

\begin{figure}[!t]
\centering
\resizebox{0.9\columnwidth}{!}{%
    \includegraphics{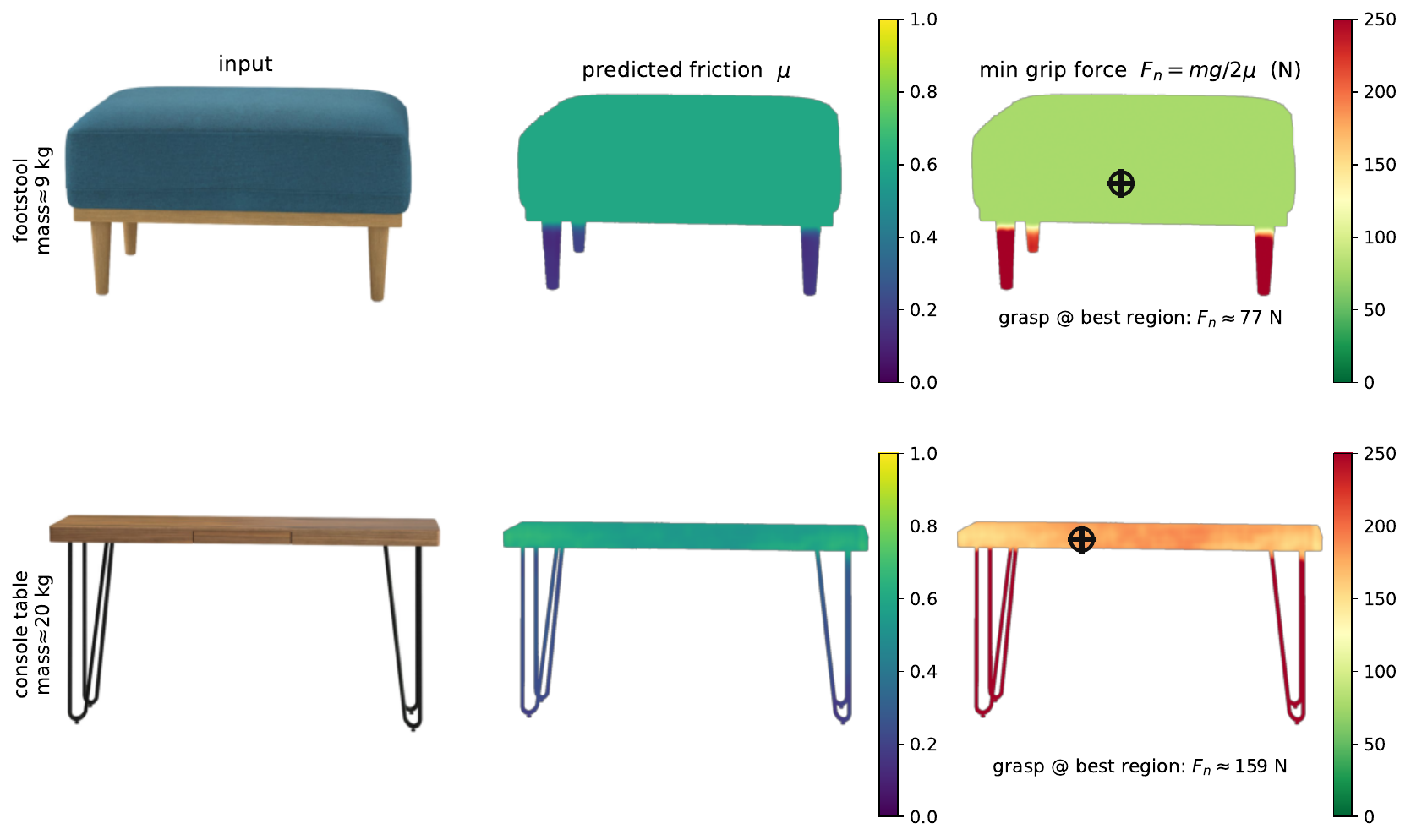}%
}
\caption{{Grasp-force planning from predicted physics.}
From one image PhysVGGT predicts per-pixel friction $\mu$ (middle) and object mass, yielding a dense minimum grip-force map $F_n=mg/(2\mu)$ (right; green$=$secure/low force, red$=$slippery/high force) and a best contact region (marker).
On both objects, the model marks the grippy top surface as the grasp region while the slick legs (wooden; thin metal) turn red ($>\!250$\,N).}
\label{fig:grasp}
\end{figure}

\paragraph{Downstream Application}
For an anti-slip two-finger pinch, the Coulomb condition $2\mu F_n \geq mg$ gives the minimum normal force $F_n=mg/(2\mu)$.
Since PhysVGGT predicts per-pixel friction $\mu$ and object-level mass $m$ from a single image, we can directly derive a dense grip-force map without depth sensing, force measurements, or physical interaction.
As shown in Figure~\ref{fig:grasp}, the resulting maps consistently favor high-friction surfaces while assigning substantially larger forces to slippery regions.
Thus, the predicted physical properties provide a spatially grounded signal for both \emph{where} to grasp and \emph{how hard} to squeeze, demonstrating their direct utility for downstream robotic manipulation.

\end{document}